**Predicting trucking accidents with truck drivers' safety climate perception across companies: A transfer learning approach**

Authors: Kailai Sun, Tianxiang Lan, Say Hong Kam, Yang Miang Goh*, and Yueng-Hsiang Huang

**Highlights (for peer-review)**

- A pretrain-then-fine-tune transfer learning approach in safety analytics is proposed
- This approach is superior to training models from scratch with limited data
- Models pretrained on larger and more diverse datasets are better
- We suggest that industry pool safety analytics data to develop better pretrained models

**Abstract**

There is a rising interest in using artificial intelligence (AI)-powered safety analytics to predict accidents in the trucking industry. Companies may face the practical challenge, however, of not having enough data to develop good safety analytics models. Although pretrained models may offer a solution for such companies, existing safety research using transfer learning has mostly focused on computer vision and natural language processing, rather than accident analytics. To fill the above gap, we propose a pretrain-then-fine-tune transfer learning approach to help any company leverage other companies' data to develop AI models for a more accurate prediction of accident risk. We also develop SafeNet, a deep neural network algorithm for classification tasks suitable for accident prediction. Using the safety climate survey data from seven trucking companies with different data sizes, we show that our proposed approach results in better model performance compared to training the model from scratch using only the target company's data. We also show that for the transfer learning model to be effective, the pretrained model should be developed with larger datasets from diverse sources. The trucking industry may, thus, consider pooling safety analytics data from a wide range of companies to develop pretrained models and share them within the industry for better knowledge and resource transfer. The above contributions point to the promise of advanced safety analytics to make the industry safer and more sustainable.

**List of Abbreviations**

| | |
|---|---|
| AI | Artificial intelligence |
| AIGC | AI-Generated Content |
| CNN | Convolutional Neural Network |
| CV | Computer Vision |
| DL | Deep Learning |
| DNN | Deep Neural Network |
| GP | Gaussian Process |
| LSTM | Long Short-Term Memory |
| ML | Machine Learning |
| NLP | Natural Language Processing |
| SMOTE | Synthetic Minority Oversampling Technique |
| WSH | Workplace Safety and Health |

# 1 Introduction

The trucking sector is a hotspot for workplace accidents and is over-represented in fatal crashes. In the United States, for instance, large trucks accounted for 9% of vehicles in fatal crashes in 2021 even though they take up only 5% of all registered vehicles (National Safety Council, 2023). The number of large trucks involved in fatal crashes has also been steadily increasing over the past decade, with their involvement rate per unit distance traveled rising by 57% from 2009 to 2021 (National Safety Council, 2023). Similarly, in Singapore, the transportation and storage industry is also an accident trouble spot with the workplace fatality rate rising by 58% from 2019 to 2023 despite the government imposing a heightened safety period from 2022 to 2023 with tighter enforcement and inspections (Ministry of Manpower, 2023). Trucks are particularly prone to serious accidents with multiple injuries or fatalities since they are often 20-30 times heavier than passenger cars, have a 20-40 percent longer braking distance and tend to override smaller vehicles in crashes (Insurance Institute for Highway Safety, 2023). Thus, there is a need to continue improving trucking safety.

Safety climate is known to be an important component in safety management. It refers to the shared perception among workers regarding their organization's policies, procedures, and practices with respect to the relative value and importance of safety (Zohar, 1980). It is a predictor of safety behavior, performance and outcomes, especially in the trucking industry (Huang et al., 2021; Kao et al., 2021; Lee et al., 2019). Furthermore, research has demonstrated that a positive safety climate in the trucking industry is linked to higher job satisfaction among truck drivers and reduced turnover rates (Huang et al., 2016). These findings underscore the critical importance of strengthening safety climate in the trucking sector.

With the rise in popularity of machine learning (ML) analysis, safety analytics has become increasingly important in the safety sciences (Sarkar & Maiti, 2020). It refers to advanced quantitative methods used to predict accidents and identify the causes and drivers of workplace safety and health (WSH) incidents, thus identifying high-risk activities to facilitate targeted interventions. Given the close relationship between safety climate and safety outcomes and performance (Huang et al., 2017), safety climate data may play an important role in accident prediction with safety analytics. Other types of safety data that can be used for such predictions include incident records, near misses, hazard reports, workers' medical information, inspection records, equipment and sensor data, and so on. Despite the versatility of safety analytics, it is constrained by the quantity (and quality) of data available since ML-based prediction usually require a large sample size (Rajput et al., 2023). This may then become a barrier for smaller organizations to use such methods since they may not have sufficient data to build artificial intelligence (AI) models in the first place. Small sample size makes it challenging to develop good ML and deep learning (DL) models and results in susceptibility to overfitting (Dos Santos et al., 2009; Kohavi & Sommerfield, 1995). Transfer learning has received increasing attention from safety sciences in recent years, though studies on using transfer learning across companies to predict workers' behaviour-related accidents are still lacking (Leoni et al., 2024).

In the current literature, studies on transfer learning and pretrained models tend to focus on computer vision (CV) and natural language processing (NLP), such as BERT model (Devlin et al., 2018) and segment anything model (Kirillov et al., 2023). Such methods are widely used to transfer knowledge from one task to another. In contrast, transfer learning and pretrained models have not been widely explored in safety analytics. Three challenges may explain this shortfall: dataset, data type, and specific DL models. In terms of dataset, developing a pretrained model to enable transfer learning typically requires a large amount of data. However, unlike CV and NLP, there are limited publicly available safety datasets due to the sensitivity of data. Second, CV and NLP usually use image or free text datatype, but safety datasets are typically tabular data. This means existing CV and NLP pretrained models cannot be directly used for transfer learning in safety analytics. Third, and related to the above, there is a lack of pretrained models (e.g., deep neural network and recurrent neural network), especially DL models, for safety analytics specifically.

Thus, to address the knowledge gap, this study aims to develop a transfer learning approach for a "target company" to utilize knowledge generated from the data of other companies or "source companies". We do so in the context of predicting truck drivers' safety outcomes with their safety climate perceptions.

The approach we propose involves pretraining AI models with data from the source companies and then fine-tuning models with data from the target company. Fine-tuning refers to using the parameters of the pretrained model as a starting point and then updating the parameters of the model when training it with data from the target company. The aim is to eventually produce a model suitable enough for the latter to predict accidents.

The main contributions of this study are as follows. (1) We introduce the pretrain-then-fine-tune approach for the above purpose and apply it in the context of safety analytics, specifically trucking accident prediction. (2) We design a deep neural network, SafeNet, to predict trucking accidents using safety climate tabular data collected from truck drivers. (3) We design three new evaluation indicators to compare the difference in performance between models trained from scratch using target company's data and models developed using our pretrain-then-fine-tune approach. (4) We compare the effect of developing pretrained models with data from different combinations of source companies. The codes are shared at https://github.com/ NUS-DBE/Pretrain-Fine-tune-safety-climate/.

## 2 Literature Review

### 2.1 Safety climate

Safety climate has been found to be an important predictor of safety performance and behavior in various industries (Christian et al., 2009; Huang et al., 2017; Nahrgang et al., 2011). A positive safety climate is especially important for the trucking sector. Truck drivers may experience isolation for extended periods of time, make real-time safety-related decisions without direct supervision, and react to emergencies without the assistance of colleagues (Huang et al., 2013; Lee et al., 2019). It is crucial to ensure that safety performance is maintained when drivers are unaccompanied, pointing to the need for a strong safety climate to minimize accidents (Lee et al., 2019). Huang et al. (2013) developed a multi-level safety climate instrument specifically for the trucking industry. It has since then been extensively used for measuring safety climate. Data collected with this instrument has been used to identify predictors of safety climate perception based on the leader-member exchange theory (Huang et al., 2021), factors of organizational commitment to safety climate (He et al., 2022) and so on. Recently, there has also been a rising research interest in analyzing safety climate perception data in the trucking industry with AI methods, such as clustering truck drivers based on their safety climate perception to recommend interventions for different driver groups (Sun et al., 2024). There is sustained interest in studying safety climate in the trucking industry.

### 2.2 Transfer learning

With the rise of using AI for safety research, transfer learning has received increasing academic attention in this field. Existing transfer learning-based safety research tends to focus on the prediction and monitoring of equipment safety rather than WSH accidents per se. For example, in a literature review on the use of AI for safety research, researchers found that the emerging interest in transfer learning in this area largely focuses on equipment fault detection (Leoni et al., 2024). Other topics studied include using computer vision to detect personal protective equipment on workers, or human and equipment detection in general (see Hung & Su, 2021; Kim et al., 2018; Lee & Lee, 2023; Tang et al., 2022). The study most relevant to the prediction of WSH accidents with transfer learning used NLP to label accidents based on free-text descriptions with non-industry-specific data and applied it to data from the construction and mining & metal industry (Goldberg, 2022). Yet even for that study, the application of transfer learning still differed from accident predictions. Thus, despite the interest in transfer learning-based safety research, insights drawn from existing studies cannot be assumed generalizable without qualification to the prediction of WSH accidents. This is usually done with behavior, perception, safety management, inspection or other project-related data (Poh et al., 2018; Sun et al., 2024). Unlike equipment fault detection, CV or NLP, accidents involving humans tend to be heavily affected by psychological and social factors, and their relationships with accidents may be less stable across contexts. This points to the need to investigate whether and how transfer learning may be used for effective accident predictions, questions that this study seeks to answer.

Transfer learning can be broadly categorized into four approaches shown in Table 1 (Niu et al., 2020; Pan & Yang, 2009; Zhuang et al., 2021). Instance transfer learning can be applied when certain parts of the data can be reused together with a few labelled data in the target domain. Studies have developed measures to identify the part of the data to be merged with the target domain's data. For example, researchers have used methods like AdaBoost to iteratively re-weigh the source domain data and assign higher weight to the correctly classified instances. Source instances not correctly classified eventually converge to zero and are not used in the final classification (Dai et al., 2007). Recent innovation in this field has incorporated attention mechanisms into a deep transfer learning model and significantly improved its performance (Wang et al., 2023). Different from the former approach which selects and reweights instances, the latter generates new samples for the target training dataset by using the source dataset with an attention mechanism.

Table 1 Transfer learning approaches

| **Transfer learning approaches** | **Description** |
|---|---|
| Instance transfer | The model is trained on instances from source domain and the knowledge is transferred at the instance level to the target domain |
| Feature representation transfer | Adapting learned feature from source domain to improve performance on a target domain |
| Parameter transfer | Transfer common parameters, i.e. priors and weights from source domain to target domain |
| Relational knowledge transfer | Transfer relational knowledge, i.e. associations, dependencies, or patterns from source domain to target domain |

As the name implies, feature representation transfer aims to find useful features and transfer them from the source domain to target domain. It is built on the basis that similar tasks may share common underlying representation. The study by Argyriou et al. (2006) is representative of early studies of feature representation transfer. It identified useful feature representation through learning a low dimensional representation shared across multiple related tasks. Today, this method is widely used in the computer vision domain. For example, studies have developed feature representation transfer learning for human activity recognition using an ensemble CNN-LSTM model (Mutegeki & Han, 2019) and proposed a transfer framework for a facial recognition system to resolve the under-represented data issues (Yin et al., 2019). Overall, feature representation transfer reduces divergence in performance when applied in different domains and results in better performance in classification and regression tasks (Zhao et al., 2024).

Similar to the previous two types of transfer learning, parameter transfer also assumes that individual models for related tasks should share some parameters or prior distribution of hyperparameters. Early parameter transfer was achieved through multi-task learning and slowly progressed into continual learning and adapter modules. One example of parameter learning through multi-task learning is a study by Lawrence and Platt (2004) who proposed MT-IVM algorithm which learns parameters of a Gaussian Process (GP) over multiple tasks by sharing the parameters. Later, continual learning is achieved through progressive neural networks proposed by Google DeepMind (Rusu et al., 2016). Continual learning is to transfer the knowledge gained from previous tasks to subsequent tasks through updating and adapting the parameters as it learns (Rusu et al., 2016). Due to the nature of stream training, networks tend to forget previous tasks after retraining, losing part of the transferred parameters. The adapter module is capable of freezing the shared parameter and thus remembering the previous task parameters (Houlsby et al., 2019).

Different from the other three approaches, the relational knowledge transfer approach transfers the relational knowledge between entities or concepts from one task to another. The goal is to leverage the understanding of relationships acquired in a source task to improve the target task. Statistical relational learning techniques are commonly used in this setting to discover relationships between datasets (Pan & Yang, 2009). One representative of this approach is TAMAR algorithm that utilizes Markov Logic Networks (MLNs) to transfer relational knowledge across different domains. TAMAR is capable of automatically mapping a source MLN to the target domain and revising its structure to further enhance its performance (Mihalkova et al., 2007).

In the current study, the proposed pretrain-then-fine-tune approach is developed based on parameter transfer. Specifically, pretraining is conducted on the source domain, while fine-tuning is conducted on the target domain, consistent with the existing parameter transfer methods.

## 3 Methodology

### 3.1 Data Source

In this study, we use the safety climate survey data collected by Huang et al. (2013). The survey uses the safety climate instrument for trucking that Huang et al. developed based on a previous study (Zohar & Luria, 2005). The survey was distributed to drivers in eight trucking companies in the U.S. through both a web-based portal and pen-and-paper method. The data collection spanned from 2011 to 2013 with 7,474 data points collected. To ensure accurate model training and evaluation, 398 incomplete data points with over 10% missing values were discarded, following Kang's (2013) recommendation. The missing values of data points with less than 10% missing values were imputed using the median replacement function suggested by Lynch (2007).

The dataset used for this study includes 42 safety climate perception items measured on a five-point Likert scale collected in the Huang et al. (2013) study, as shown in Table 2. Other than the safety climate perception items, the survey also asked drivers the number of accidents they suffered in the past year. We converted answers to this question into a binary variable of whether one had accidents in the past year. This variable is the target in the AI models in this study, while the 42 safety climate items are the features.

Table 2 Safety climate perception items used in this study

| S/N | Description |
|---|---|
| 1 | Uses any available information to improve existing safety rules |
| 2 | Tries to continually improve safety levels in each department |
| 3 | Invests a lot in safety training for workers |
| 4 | Creates programs to improve drivers' health and wellness (e.g., diet, exercise) |
| 5 | Listens carefully to our ideas about improving safety |
| 6 | Cares more about my safety than on-time delivery |
| 7 | Allows drivers to change their schedules when they are getting too tired |
| 8 | Provides enough hands-on training to help new drivers be safe |
| 9 | Gives safety a higher priority compared to other truck companies |
| 10 | Reacts quickly to solve the problem when told about safety concerns |
| 11 | Is strict about working safely when delivery falls behind schedule |
| 12 | Gives drivers enough time to deliver loads safely |
| 13 | Fixes truck/equipment problems in a timely manner |
| 14 | Will overlook log discrepancies if I deliver on time |
| 15 | Makes it clear that, regardless of safety, I must pick up/deliver on time |
| 16 | Expects me to sometimes bend safety rules for important customers |
| 17 | Turns a blind eye when we use hand-held cell phones while driving |
| 18 | Assigns too many drivers to each supervisor, making it hard for us to get help |
| 19 | Hires supervisors who don't care about drivers |
| 20 | Turns a blind eye when a supervisor bends some safety rules |
| 21 | Compliments employees who pay special attention to safety |
| 22 | Provides me with feedback to improve my safety performance |
| 23 | Respects me as a professional driver |
| 24 | Frequently talks about safety issues throughout the work week |
| 25 | Discusses with us how to improve safety |
| 26 | Uses explanations (not just compliance) to get us to act safely |
| 27 | Is supportive if I ask for help with personal problems or issue |

| S/N | Description |
|---|---|
| 28 | Is an effective mediator/trouble-shooter between the customer and me |
| 29 | Is strict about working safely even when we are tired or stressed |
| 30 | Gives higher priority to my safety than on-time delivery |
| 31 | Would like me to take care of serious equipment problems first before delivering |
| 32 | Gives me the freedom to change my schedule when I see safety problems |
| 33 | Makes me feel like I'm bothering him/her when I call |
| 34 | Encourages us to go faster when deadheading (going for a new load) |
| 35 | Expects me to sometimes bend driving safety rules for important customers |
| 36 | Sometimes turns a blind eye with rules when deliveries fall behind schedule |
| 37 | Pushes me to keep driving even when I call in to say I feel too sick or tired |
| 38 | Expects me to answer the cell phone even while I'm driving |
| 39 | Stops talking to me on the phone if he/she hears that I am driving |
| 40 | Turns a blind eye when we use hand-held cell phones while driving |
| 41 | Is an effective mediator / trouble-shooter between management and me |
| 42 | Regularly asks me if I have had enough sleep |

### 3.2 Data Imbalance

Table 3 Sample sizes of data from different companies

| Company | Sample Size | Balanced sample size (after SMOTE) | No. of samples with accident(s) | Test set | Training set |
|---|---|---|---|---|---|
| **1** | 1831 | 2178 | 742 | 200 | 0-1978 |
| **2** | 432 | 446 | 209 | 178 | 0-268 |
| **3** | 257 | 460 | 27 | 184 | 0-276 |
| **4** | 244 | 396 | 46 | 158 | 0-238 |
| **5** | 218 | 320 | 58 | 128 | 0-192 |
| **6** | 3493 | 5426 | 780 | 200 | 0-5226 |
| **7** | 406 | 588 | 112 | 200 | 0-388 |

Although the survey was conducted with eight companies, the drivers from one of the companies did not report any accidents. Thus, we only used data from the other seven companies. This dataset has two imbalances: one is the imbalanced sample size for different companies (company imbalance), the other is the imbalanced sample size of the two classes in the target (class imbalance). For company imbalance, we can see that more samples are from companies 1 and 6 and fewer samples are from companies 2, 3, 4, 5 and 7, as shown in Table 3. This imbalance is advantageous to this study since it creates the variety in dataset sizes of the target company, allowing the transfer of insights derived from larger source companies' data to smaller target companies. Consider, for example, if company 5 with only 218 samples wants to develop an AI model to predict accidents from safety climate data, results from this study may help it to do so by transferring knowledge gained from data from other larger companies to this small company.

As highlighted, all seven datasets have class imbalance, i.e., the sample size of drivers with one or more accidents is smaller than the sample size of drivers without accident. Class imbalance is a very important issue since it may bias the AI models whereby the latter tends to favor the majority class while neglecting the minority class. To address this, we use SMOTE method (Chawla et al., 2002) to generate new datapoints from the minority class. This is done by randomly selecting a minority class sample and then randomly choosing one of its nearest neighbors in the minority class. The synthetic sample is then created by interpolating between these two existing samples. This approach effectively increases the number of instances of the minority class in the dataset, helping to balance the class distribution.

### 3.3 Pretraining and fine-tuning pipeline

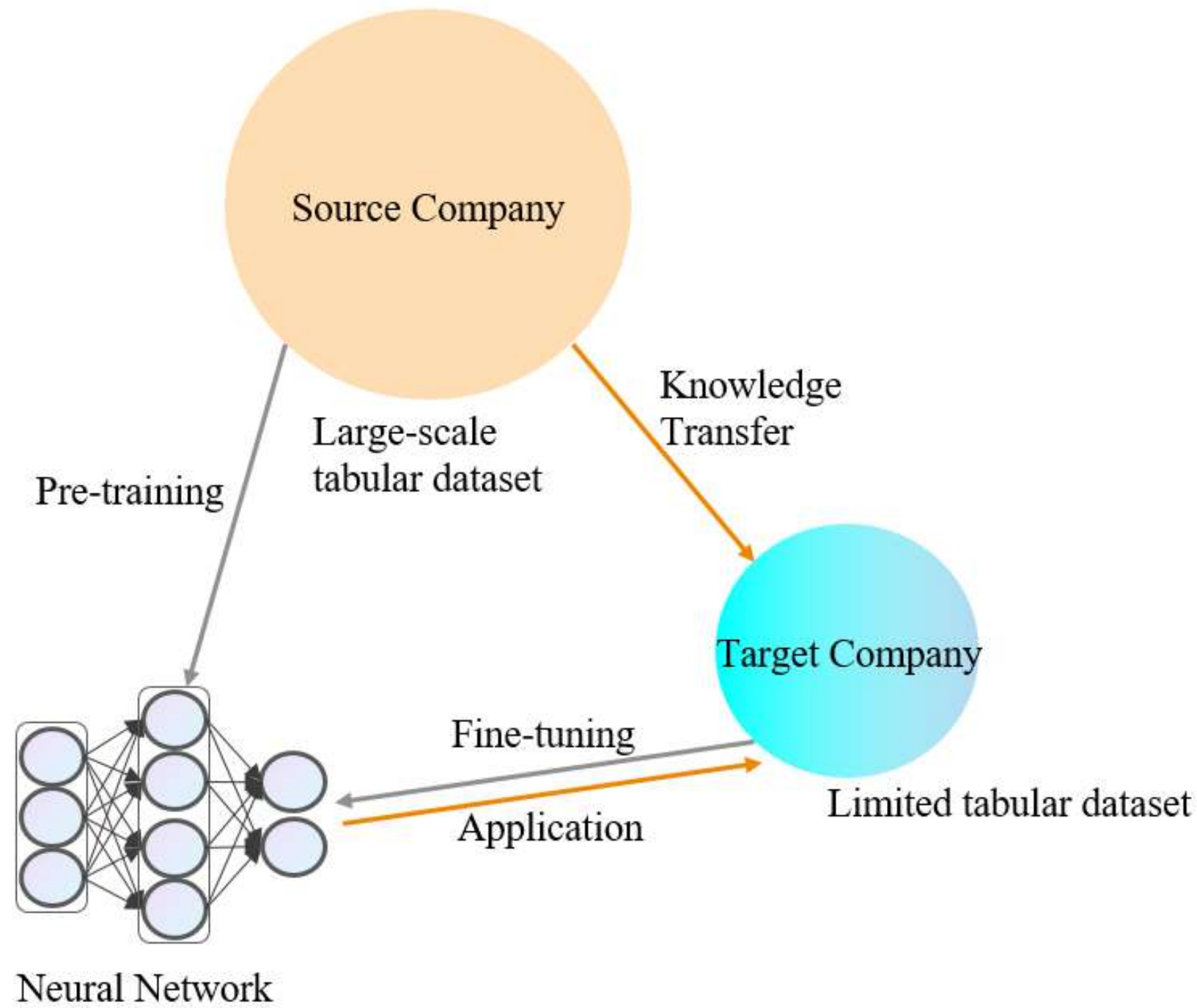

Figure 1 Illustration of pretraining and fine-tuning pipeline

Figure 1 illustrates the pretrain-then-fine-tune approach we propose in this paper to help any target company develop more accurate AI models using the knowledge generated from data from source companies. We first pretrain a deep neural network (DNN) model using the data from the source companies. Both the training set and testing set are from source companies, and a 70/30 split is used for training and testing, respectively. Given $N$ sample pairs $\{\boldsymbol{x}_s^n, y_s^n\}$, we train a DNN model by minimizing the cross-entropy loss function. $\boldsymbol{x}$ represents the feature vector, and $y$ represents the label (accident or not). $n$ represents the $n_{th}$ sample; $s$ represents that the data sample is from source companies; $k$ represents the class number. This task aims to learn a function $f_\theta(.)$ to map the feature $\boldsymbol{x}_s^n$ to the label $y_s^n$. Then, we choose the best-accuracy model as the pretrained model. We define the best-accuracy model parameters as $\theta_s$. The training process can be formulated as:

$$\theta_s = argmin\ \frac{1}{N}\sum_{n=1}^{N}\sum_{k=0}^{K} -\mathrm{y}_{s,k}^n \log f_\theta(\boldsymbol{x}_s^n). \tag{1}$$

Second, we fine-tune the pretrained model using the data from the target company:

$$\theta_{pt} = argmin\ \frac{1}{M}\sum_{m=1}^{M}\sum_{k=0}^{K} -\mathrm{y}_{t,k}^m \log f_{\theta|\theta_s}(\boldsymbol{x}_t^m) \tag{2}$$

where $t$ represents that the data sample is from target company, and $M$ represents the training sample size of target company. $f_\theta$ refers to a model trained from scratch with the same structure as the pretrained model using the target company data. $f_{\theta|\theta_s}$ refers to the model updated based on the pretrained parameters after fine-tuning with the target company data. This is relevant to the real-life scenario. A target company may have limited data compared to source companies. If the former uses the limited data to train their models from scratch, $f_{\theta|\theta_s}$ is equal to $f_\theta$. What we suggest is to let the target company to use data from source companies to pretrain the model, and then use the target company's own data to update (i.e., fine-tune) the parameters from the basic parameters $f_{\theta|\theta_s}$ of the pretrained model. Thus, the overall optimization problem can be formulated as:

$$min\ \frac{1}{M}\sum_{m=1}^{M}\sum_{k=0}^{K} -\mathrm{y}_{t,k}^{m} \log f_{\theta|\theta_s}(\boldsymbol{x}_t^m)\,,$$
$$\text{s.t.} \begin{cases} SCD \cap TCD = \emptyset, \\ M < m_0, \end{cases} \tag{3}$$

where SCD refers to source company data and TCD refers to target company data. The data in SCD and TCD should be different, i.e., $SCD \cap TCD = \emptyset$. It is because, in practice, it is impossible that the pretrained model has "seen" the data from target companies. Another constraint is that the sample size of TCD should be smaller than a parameter $m_0$ to fit the practical conditions.

### 3.4 Dataset preparation

To analyze the effectiveness of the proposed pretrain-then-fine-tune approach, we need to implement the DNN model training and evaluation process. A key step lies in the dataset preparation. To ensure that our proposed approach is tested in diverse scenarios, we constructed the datasets with the different combinations of the seven companies' data used in this study, as shown in Table 4.

Table 4 Source and target datasets

| S/N | Source dataset (for pretraining) | Target dataset (for fine-tuning) | Type |
|---|---|---|---|
| 1 | Company 3 | Company 3 | One-to-one |
| 2 | Company 3 | Company 4 | |
| 3 | Company 3 | Company 5 | |
| 4 | Company 3 | Company 6 | |
| 5 | Company 3 | Company 7 | |
| 6 | Company 4 | Company 3 | |
| 7 | Company 4 | Company 4 | |
| 8 | Company 4 | Company 5 | |
| 9 | Company 4 | Company 6 | |
| 10 | Company 4 | Company 7 | |
| 11 | Company 5 | Company 3 | |
| 12 | Company 5 | Company 4 | |
| 13 | Company 5 | Company 5 | |
| 14 | Company 5 | Company 6 | |
| 15 | Company 5 | Company 7 | |
| 16 | Company 6 | Company 3 | |
| 17 | Company 6 | Company 4 | |
| 18 | Company 6 | Company 5 | |
| 19 | Company 6 | Company 6 | |
| 20 | Company 6 | Company 7 | |
| 21 | Company 4+5 | Company 3 | Two-to-one |
| 22 | Company 4+5 | Company 4 | |
| 23 | Company 4+5 | Company 5 | |
| 24 | Company 4+5 | Company 6 | |
| 25 | Company 4+5 | Company 7 | |
| 26 | Company 6 | Company 7 | More-to-one |
| 27 | Company 1+6 | Company 7 | |
| 28 | Company 1+2+6 | Company 7 | |
| 29 | Company 1+2+3+6 | Company 7 | |
| 30 | Company 1+2+3+4+5+6 | Company 7 | |

In the pretraining process, we split the source dataset into 70/30 for training and testing. We also split the target dataset for fine-tuning into training and testing sets. To show the effectiveness of our proposed approach, we conducted the fine-tuning process at various sizes of the training set, ranging from 1 to 60% of the total data points in the target dataset. For example, as shown in Table 3, for company 2, we conducted the fine-tuning with the size of training set ranging from 1 to 268 ($0.6 * 446 \approx 268$). Meanwhile, the testing set is fixed at 40% of the total data points in the target dataset, regardless of the number of training samples. For example, as shown in Table 3, for company 2, $0.4 * 446 \approx 178$ is the size of the testing set.

After dataset splitting, for each training and testing, we normalized the feature vector $\boldsymbol{x}^i$:

$$x_j^i = \frac{x_j^i - \mu_j}{\sigma_j}, \quad (4)$$

where $x_j^i$ represents the $i_{th}$sample and the $j_{th}$ feature, and $\mu_j$ represents the mean of the $j_{th}$ feature, and $\sigma_j$ is the standard deviation of the $j_{th}$ feature.

### 3.5 Deep neural network: SafeNet

To feed the safety climate dataset, we designed a custom deep neural network in Figure 2, named SafeNet. This model is architecturally designed for classification tasks, with a primary focus on accident prediction based on various input features. The SafeNet comprises several linear layers, with the input layer accepting a feature set of size 42. The architecture progresses through fully connected transformations, escalating from the input size to 64, and then to 128.

Key attributes of the model include ReLU activation functions to introduce non-linearity, batch normalization for enhanced training stability and efficiency. We use skip connections to prevent gradient vanishing and explosion in deep neural networks. Skip connections allow gradients to flow directly through the network by adding the input of a layer to its output. This facilitates network training by mitigating the vanishing gradient problem and helps to control gradient explosion. A notable feature is the incorporation of a residual connection between the second and third fully connected layers, which is instrumental in learning complex data patterns more effectively. The concluding part of the model is a fully connected layer that narrows the output to two dimensions, suitable for binary classification objectives. We chose this architecture to ensure a potent and efficient learning mechanism, particularly for applications in safety analytics and evaluation and prediction.

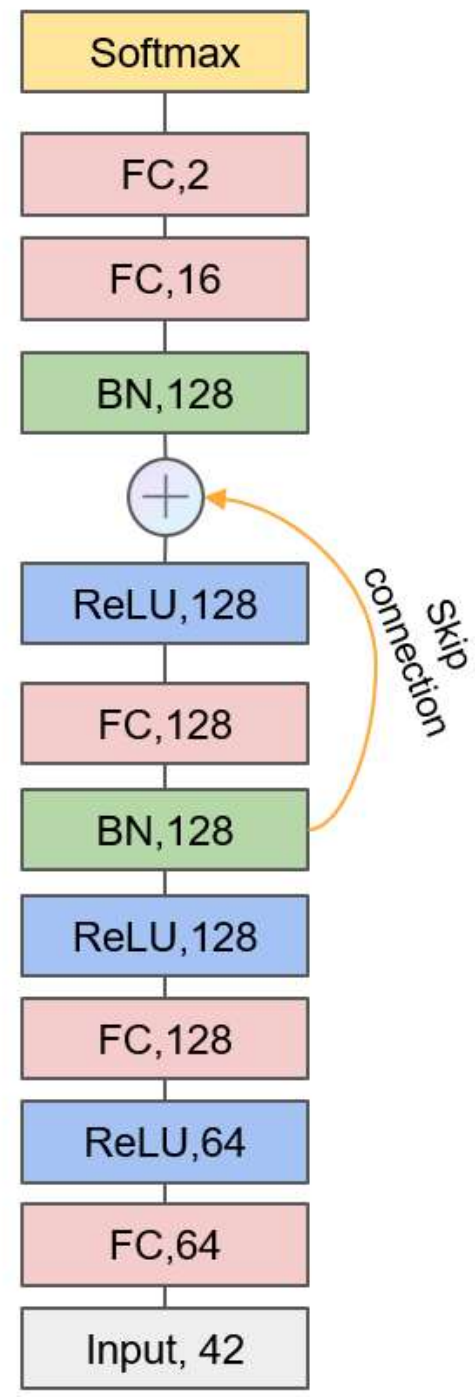

Figure 2 SafeNet structure

### 3.6 Hyper-parameter

Choosing hyper-parameters is important for training SafeNet. It is divided into two parts: pretraining and fine-tuning. (1) In the pretraining step, we used AdamW optimizer, a variant of the traditional Adam optimizer known for its effectiveness in handling weight decay. The optimizer was configured with a learning rate of 0.0003, tailored to balance the rate of convergence and training stability. We set weight_decay as 0.01 for regularization to prevent overfitting. To further enhance the robustness during the training process, we integrated a learning rate scheduler, specifically the Step Learning Rate (StepLR) scheduler. This scheduler was programmed to reduce the learning rate by a factor of 0.1 every 30 epochs. The rationale behind this approach was to adaptively decrease the learning rate as the training progressed, allowing for finer adjustments to the model parameters in the later stages of training. This strategy is conducive for the model to converge to a more optimal one. We trained our SafeNet at 100 epochs. (2) In the fine-tuning step, we trained models using a 0.001 learning rate during 20 epochs. Batchsize was set to 4.

### 3.7 Evaluation indictors

After using SMOTE oversampling method, we obtained a balanced dataset. We used standard accuracy to evaluate our method on the testing dataset:

$$A = \frac{1}{Z}\sum_{z=1}^{Z}\sum_{k=0}^{K}(1 - \mathrm{sign}(y_k^z - \hat{y}_k^z)), \tag{5}$$

where the $\hat{y}_k^z$ represents the predicted accident, and $y_k^z$ represents the label. $Z$ is the size of the testing dataset.

During the fine-tuning stage, we trained two models on the target dataset. First, we trained SafeNet using this dataset alone with the model's parameters $\theta$ initialized from a standard normal distribution. The resulting parameters after training were $\theta_{st}$. Second, we trained SafeNet based on the model pretrained using source dataset with parameters $\theta_s$, and the parameters $\theta$ were initialized from $\theta_s$. The resulting parameters after training were $\theta_{pt}$. In lay terms, the key difference between the two models is that one was developed using the pretrained model as the starting point, the other was developed from scratch using target dataset alone. To highlight the benefits of pretraining, we compared the difference in accuracy (DA) of the two models:

$$\mathrm{DA} = A_{\theta_{pt}} - A_{\theta_{st}} = \frac{1}{Z}\sum_{z=1}^{Z}\sum_{k=0}^{K}(\mathrm{sign}(\mathrm{y}_k^z - \hat{\mathrm{y}}_{k,\theta_{st}}^z) - \mathrm{sign}\left(\mathrm{y}_k^z - \hat{\mathrm{y}}_{k,\theta_{pt}}^z\right)), \tag{6}$$

Given that for each set of target dataset we used a wide range of sizes for the training set as shown in Table 3, we used the DA values across different training set sizes to calculate effective proportion (EP). In lay terms, EP means, across the different training set sizes tested, the proportion of times when our pretrain-then-fine-tune method resulted in higher accuracy than developing the model from scratch using the target dataset alone. We used $S$ to represent the number of times the training (i.e., fine-tuning) was conducted across different training set sizes. We obtained:

$$\mathrm{EP} = \frac{1}{S}\sum_{s=1}^{S} sign\ (A_{\theta_{pt}}^{s} - A_{\theta_{st}}^{s}). \tag{7}$$

Similarly, we also computed the degree to which $A_{\theta_{pt}}^{s}$ is greater than $A_{\theta_{st}}^{s}$:

$$ME = \frac{1}{S}\sum_{s=1}^{S}\left(A_{\theta_{pt}}^{s} - A_{\theta_{st}}^{s}\right). \tag{8}$$

$$NME = \frac{1}{S}\sum_{s=1}^{S}\frac{A_{\theta_{pt}}^{s} - A_{\theta_{st}}^{s}}{A_{\theta_{st}}^{s}}. \tag{9}$$

ME means the mean error between the accuracy of our pretrain-then-fine-tune method and the accuracy of developing the model from scratch using the target dataset alone, across the different training set sizes tested. NME means the normalized degree of ME.

## 4 Results and Discussion

As explained in Table 4, we conducted many experiments on different datasets from different companies. Section 4.1 aims to answer: is pretraining effective on safety climate tabular data? Sections 4.2 and 4.3 aim to answer: does pretraining with data from more companies improve the performance of fine-tuned models? In section 4.1, we explain the result of pretraining with SafeNet on one company's data and then fine-tuning the model on another company's data, i.e., one-to-one transfer. Besides, we also explain how the target datasets for fine-tuning are selected, to provide a fair and complete picture of the results from this study. Using the example of company 2 as a set of "error cases", we also illustrate the importance of pretraining accuracy to our proposed pretrain-then-fine-tune approach. Section 4.2 shows the result of pretraining with SafeNet on two companies' data and then fine-tuning the model on one company's data, i.e., two-to-one transfer. Section 4.3 shows the result of pretraining on many companies' data and fine-tuning on one company' data, i.e., more-to-one transfer.

## 4.1 One-to-one

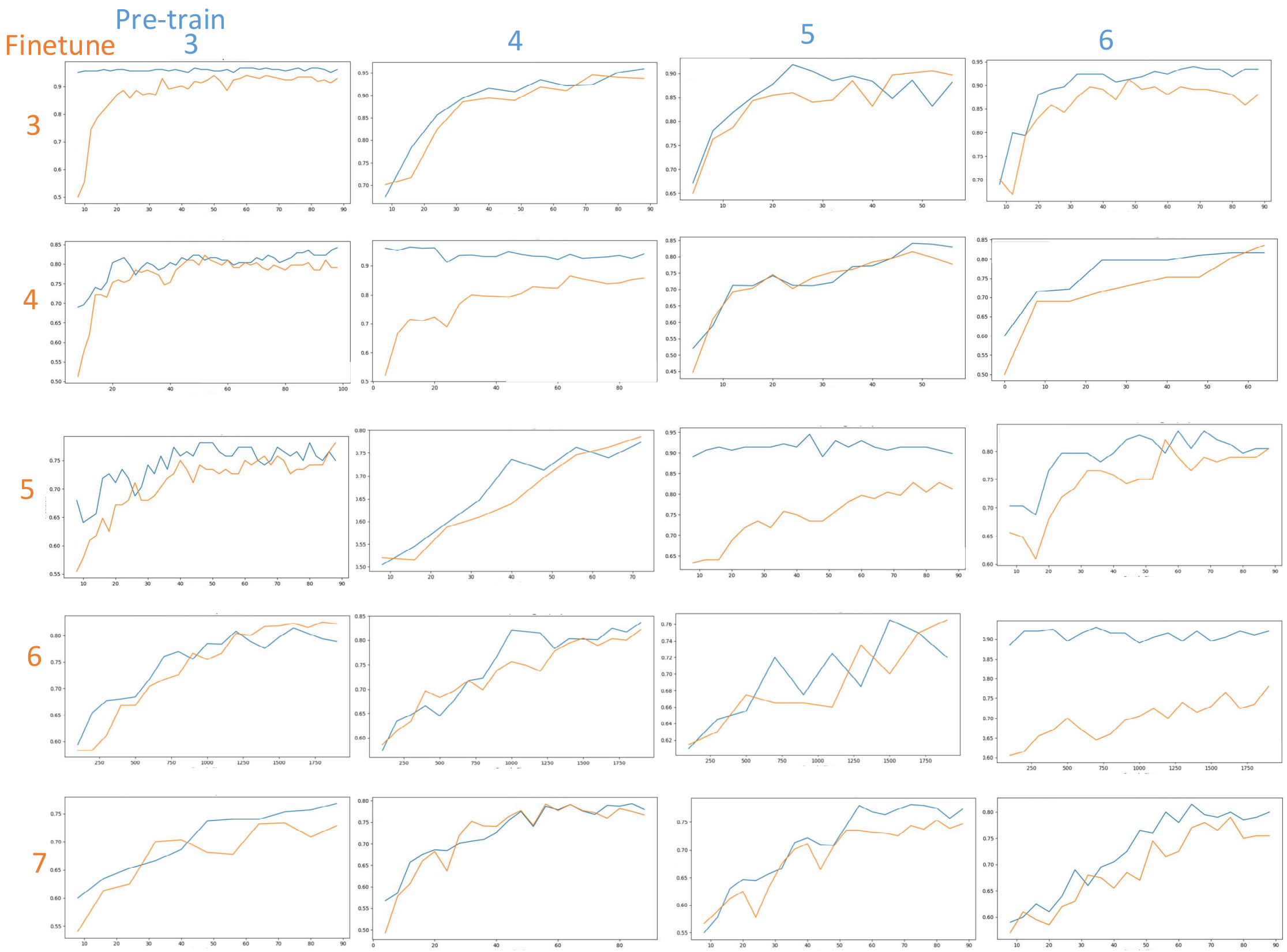

Figure 3 Accident prediction accuracy with and without pre-training for one-to-one transfer

*The numbers (3, 4, 5, 6, 7) represent company ID. Blue line is the accuracy of the model pretrained on the source dataset and fine-tuned on the target dataset. Orange line is the accuracy of the model developed using the target dataset only without pretraining. X-axis is the sample size for each class in the training set of the target dataset, i.e., if $x$=100, then the size of the training set is 2*100=200. Y-axis is the precision.

Table 5 EP (effective proportion) values of models tested for one-to-one transfer

| | | **Source dataset (for pretraining)** | | | |
|---|---|---|---|---|---|
| | | Company 3 | Company 4 | Company 5 | Company 6 |
| **Target dataset (for fine-tuning)** | Company 3 | 1 | 0.836 | 0.732 | 0.996 |
| | Company 4 | 0.958 | 1 | 0.596 | 0.93 |
| | Company 5 | 0.91 | 0.76 | 1 | 0.97 |
| | Company 6 | 0.6 | 0.754 | 0.69 | 1 |
| | Company 7 | 0.822 | 0.498 | 0.898 | 0.932 |

Table 6 ME (mean error) values of models tested for one-to-one transfer

| | | **Source dataset (for pretraining)** | | | |
|---|---|---|---|---|---|
| | | Company 3 | Company 4 | Company 5 | Company 6 |
| **Target dataset (for fine-tuning)** | Company 3 | 7.3576 | 1.6064 | 1.3189 | 4.1599 |
| | Company 4 | 2.749 | 15.1691 | 0.6475 | 4.44859 |
| | Company 5 | 3.2538 | 2.1106 | 16.274 | 4.476 |
| | Company 6 | 0.9554 | 1.5112 | 1.33703 | 21.3186 |
| | Company 7 | 2.4357 | 0.3866 | 2.3991 | 3.3097 |

Table 7 NME (normalized degree of mean error) values of models tested for one-to-one transfer

| | | Source dataset (for pretraining) | | | |
|---|---|---|---|---|---|
| | | Company 3 | Company 4 | Company 5 | Company 6 |
| **Target dataset (for fine-tuning)** | Company 3 | 47.63335 | 9.746204 | 8.417833 | 24.69386 |
| | Company 4 | 19.2805 | 102.032 | 4.825934 | 31.85511319 |
| | Company 5 | 22.150564 | 17.27887 | 111.7672 | 30.92804 |
| | Company 6 | 8.358988 | 9.896377 | 10.182251 | 155.3554 |
| | Company 7 | 18.30381 | 3.759725 | 17.21945 | 23.75763 |

This section shows the results of one-to-one transfer models, and subsection 4.1.1 explains how the target datasets for fine-tuning are selected. Figure 3 shows the models' accuracy and Table 5 to Table 7 show the values of their EP, ME and NME. For convenience of expression, we use *P-Q* to represent pretraining the model on data from company P and fine-tuning on data from company Q. For example, 3-4 means pretraining on company 3 and fine-tuning on company 4, corresponding to the second row of first column in Figure 3. Below we summarize the insights drawn from the results.

(1) Across all graphs, model accuracy generally shows an upward trend with an increasing volume of the target dataset. This indicates that more fine-tuning data is beneficial for performance improvement. In real life, collecting more data from a target company would be helpful for developing better models.

(2) In Table 5, all EP values are greater than 0.5; in Table 6 and Table 7, all values are greater than 0. These results demonstrate that the proposed pretrain-then-fine-tune approach can achieve a better performance compared to developing the model from scratch using the target dataset.

(3) If the datasets used in pretraining are small, performance will not be high even after the model is fine-tuned on a large dataset. For example, the models pretrained with data from company 3, 4, or 5 did not perform well after fine-tuning with data from company 6, even though they performed better than models developed with data from company 6 without pretraining. In contrast, models pretrained with data from company 6 generally performed well. This means larger datasets are strongly preferred over smaller ones in pretraining. This is also consistent with the norm of pretraining on larger datasets and fine-tuning on smaller datasets in deep learning for computer vision or natural language processing. That said, if a target company cannot acquire models pretrained on large datasets, pretraining the model on smaller datasets is still better than developing the model from the target dataset without pretraining. As shown in Figure 3, for example, for models pretrained on smaller companies (3, 4, 5), the blue lines are generally above the orange lines, which means pretraining still results in better performance of the models eventually developed. This means that for **tabular datasets** like the ones used in this study, even pretraining on smaller datasets adds value to the eventual model performance.

(4) The plots along the diagonal line in Figure 3 (3-3, 4-4, 5-5) show the results when we pretrain and fine-tune the model on the same company's data, i.e., the pretraining and target datasets are identical. When fine-tuning a model using the pretrained model as the starting point, the change in performance with increasing training set size is more stable. In contrast, when training a model with target company data and no pretraining, accuracy increases from 50% as the training set size rises. This is expected, since using the same dataset for pretraining and fine-tuning should result in good performance given the pretrained model has "seen" the fine-tuning data. Nevertheless, given that company 6 has more data, for 6-6, we set the pretraining and target datasets as mutually exclusive. $\mathrm{SCD} \cap \mathrm{TCD} = \emptyset$. This is done by oversampling company 6's data with SMOTE and splitting it into two parts, pretraining on one part (60%) and fine-tuning on the other (40%). The above results are compared with models trained with the target dataset alone. The pretraining and target datasets in this case have identical distribution. The result also demonstrates the effectiveness of our pretrain-fine-tuning method.

(5) Company 6's data are relatively difficult to train, with the accuracy of its pretrained model at only 81.6%, as shown in Table 8. Despite this, the pretrained model performed well. Table 5 shows that models pretrained on company 6's data generally performed better than other models. This reiterates the benefits of our pretrain-then-fine-tune approach.

### 4.1.1 Pretraining accuracy and error cases

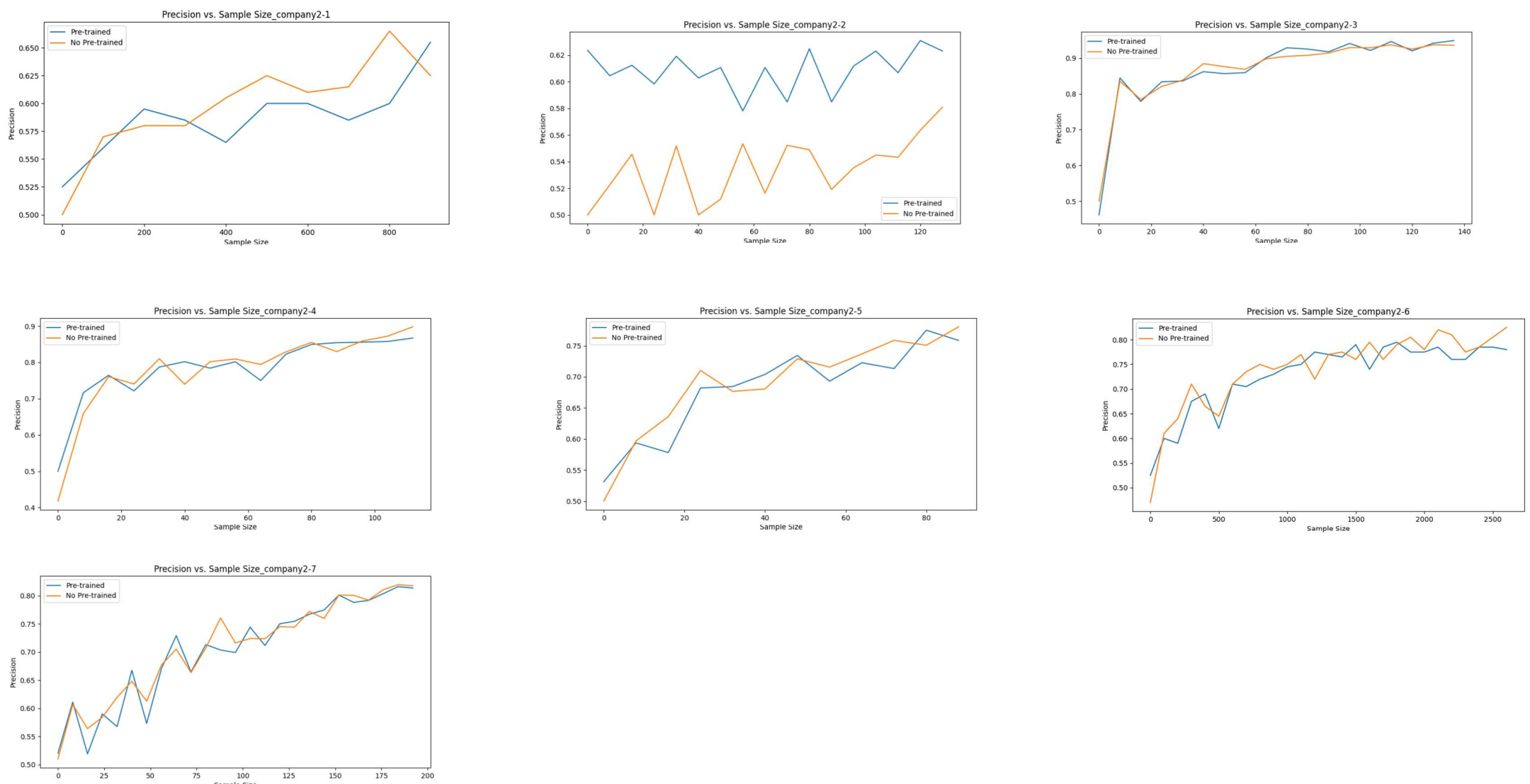

Figure 4 Results of pretraining SafeNet on data from company 2 and then fine-tuning the model on data from companies 1-7

Table 8 SafeNet accuracy after pretraining using different company's data

| **Source dataset (for pretraining)** | Company 1 | Company 2 | Company 3 | Company 4 | Company 5 | Company 6 |
|---|---|---|---|---|---|---|
| **Accuracy of SafeNet** | 0.599 | 0.567 | 0.928 | 0.916 | 0.885 | 0.816 |

As shown in Table 8, model accuracy after pretraining on data from companies 1 and 2 is lower than pretraining on data from other companies. We found that if the model accuracy is low after pretraining, the pretraining contributes little to model accuracy during the whole pretrain-then-fine-tune process. Using company 2's data to illustrate this point, Figure 4 and Table 9 show the comparison of the model performance with and without pretraining when the target dataset is from companies 1-7 respectively. Figure 4 shows that the model accuracy when pretraining with company 2's data and fine-tuning with the target company's data (blue lines) is generally lower than or similar to training with the target company's data alone without pretraining (orange lines). Similarly, in Table 9, most EP values are not greater than 0.5 and most ME/NME values are not greater than 0 (pretraining on company 2's data resulted in high EP, ME and NME because the pretrained model has "seen" fine-tuning data). This suggests that there is minimal benefit in pretraining the model on company 2's data and fine-tuning with the target company's data, compared to developing the model with the target company's data alone.

Given the poor pretraining accuracy on data from companies 1 and 2, we selected data from companies 3-7 only for one-to-one transfer, as shown in section 4.1. This makes sense theoretically since if we use low-accuracy pretrained models to fine-tune, it is less likely to transfer the correct domain knowledge. The takeaway of this subsection is that when conducting transfer learning with practical importance downstream, one should first ensure that the pretrained model achieves a good performance.

Table 9 EP (effective proportion), ME (mean error) and NME (normalized degree of mean error) values of models tested for one-to-one transfer when pretraining on company 2 data

| Target dataset (for fine-tuning) | EP | ME | NME |
|---|---|---|---|
| Company 1 | 0.306 | -1.464 | -11.4336 |
| Company 2 | 1 | 7.363 | 69.43388 |
| Company 3 | 0.582 | 0.0581 | -0.0134 |
| Company 4 | 0.382 | 0.2375 | 3.584662 |
| Company 5 | 0.384 | -0.956 | -6.62688 |
| Company 6 | 0.234 | -1.112 | -7.09046 |
| Company 7 | 0.404 | -0.594 | -4.55438 |

### 4.2 Two-to-one

This section explains the results of two-to-one transfer models. As indicated in Table 4, the source dataset is constituted by data from companies 4 and 5. Figure 5 and Table 10 show the results. One may first notice that in Table 10, all EP values are greater than 0.5 and all ME/PME values are greater than 0. These results demonstrate that the pretrain-then-fine-tune approach can achieve a better performance compared with building models from scratch with the target dataset only. Furthermore, comparing Table 10 with the columns for companies 4 and 5 from Table 5 to Table 7, it may be observed that if we merge more companies' data into the pretraining dataset, the EP/ME/NME values will improve. Thus, the takeaway is that when doing transfer learning, the target company should strive to obtain large and diverse datasets to establish a good enough pretrained model.

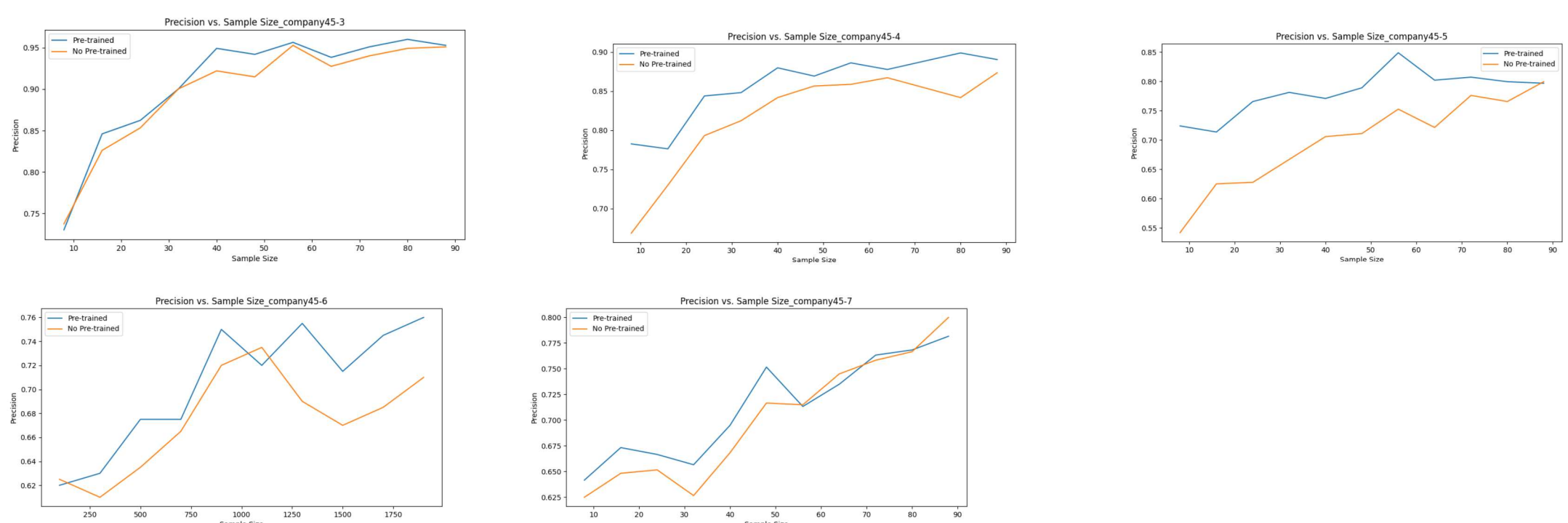

Figure 5 Accident prediction accuracy with and without pre-training for two-to-one transfer

* Blue line is the accuracy of the model pretrained on the source dataset and fine-tuned on the target dataset. Orange line is the accuracy of the model developed using the target dataset only without pretraining. X-axis is the sample size for each class in the training set of the target dataset, i.e., if *x*=100, then the size of the training set is 2*100=200. Y-axis is the precision.

Table 10 EP (effective proportion), ME (mean error) and NME (normalized degree of mean error) values of models tested for two-to-one transfer

| Target dataset (for fine-tuning) | EP | ME | NME |
|---|---|---|---|
| Company 3 | 0.972 | 1.1696049243533256 | 6.446359964019697 |
| Company 4 | 1.0 | 3.769695421711621 | 23.714103601306803 |
| Company 5 | 1.0 | 8.246296614302707 | 61.537689469276465 |
| Company 6 | 0.924 | 3.0818361509001357 | 22.80273282603771 |
| Company 7 | 0.736 | 1.2568262116494361 | 9.629764362708686 |

### 4.3 More-to-one

This section explains the results of more-to-one transfer models. As shown in Table 4, we pretrain the models by incrementally adding more companies' data into the source dataset to present an ablation study. The target dataset in this case is constituted of data from company 7. Figure 6 shows how pretraining can improve a model's precision. There are several lines on the chart, each representing a different training strategy. (1) The blue line indicates the precision of a model built on the target dataset alone without pretraining. (2) The other lines represent the precision of models pretrained on varying numbers of company data. For example, "Pretrained on company 6" (the orange line) suggests that the model was pretrained on data from company 6, and this pretraining contributes to how the model performs on data from company 7. The figure shows that as more companies' data are included in pretraining, the precision of the eventual model after fine-tuning on company 7's data improves, illustrating the benefit of using diverse pretraining data.

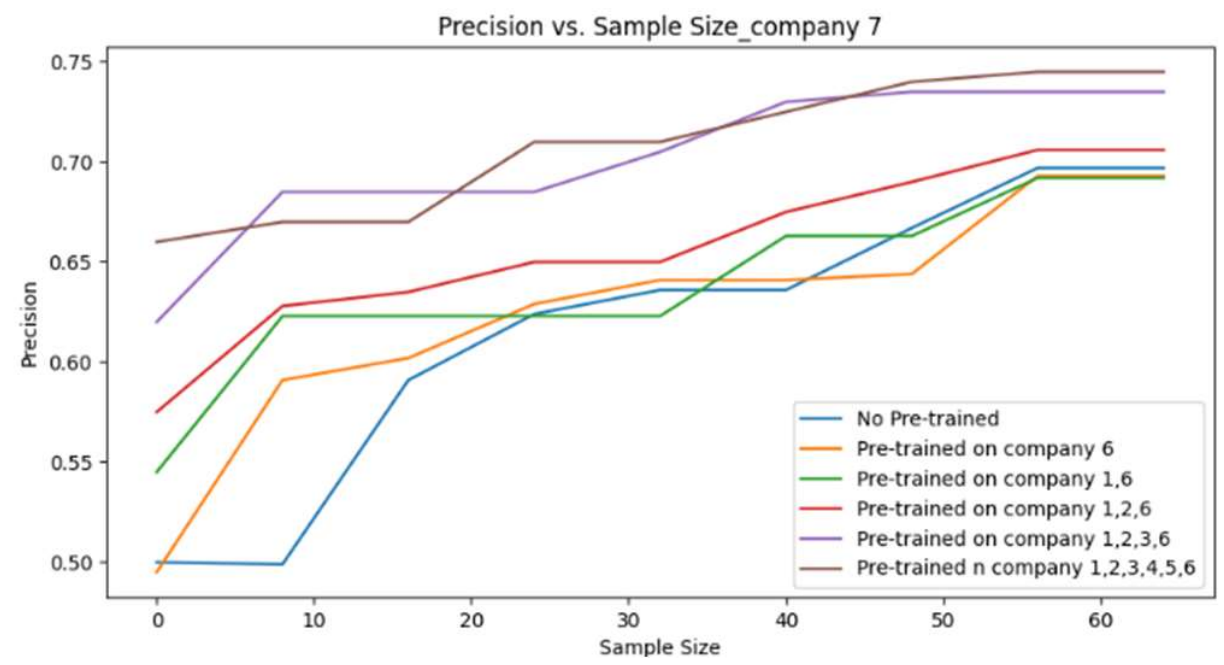

Figure 6 The performance comparison of pretraining on datasets with data from different combinations of companies with increasing sample size of the target dataset

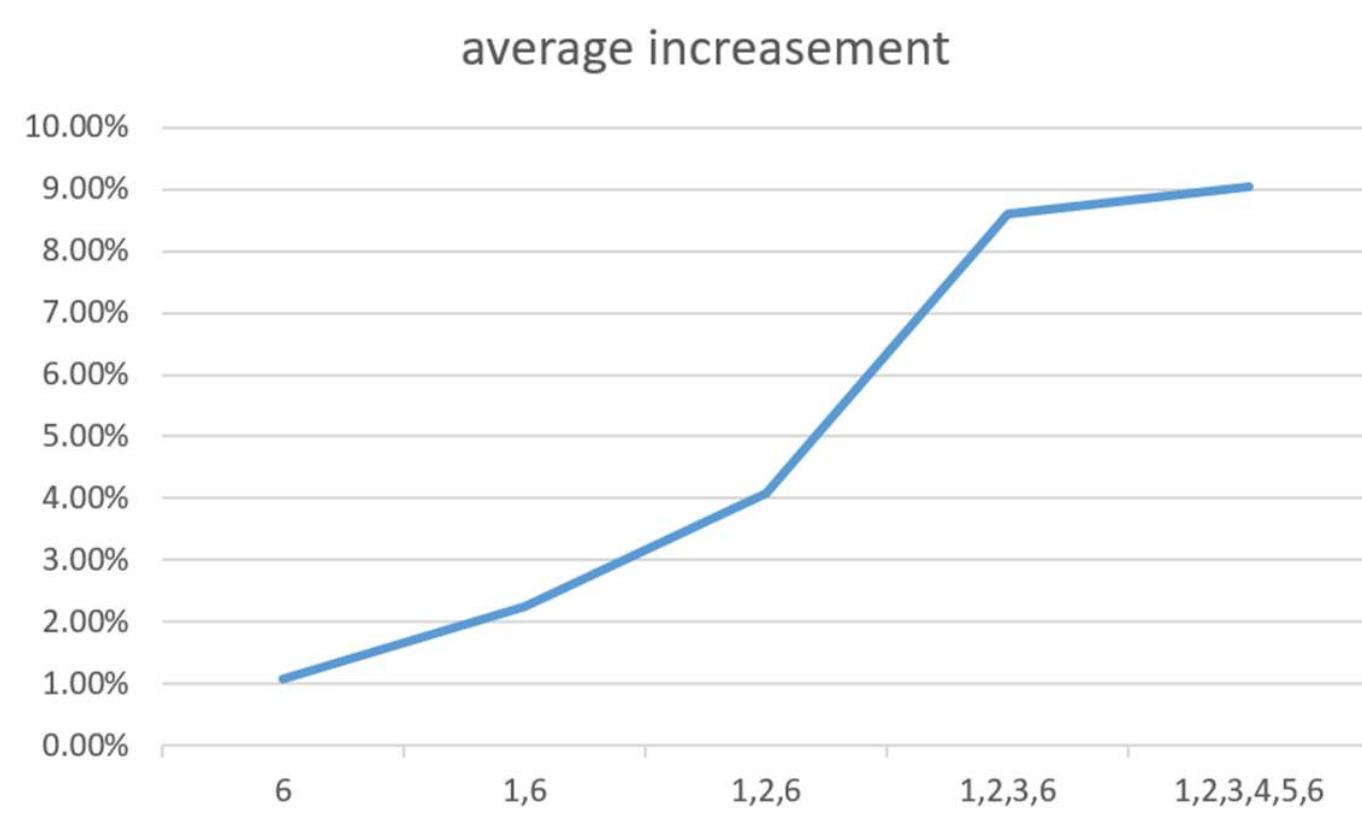

Figure 7 Average difference in fine-tuning accuracy compared to training a model from scratch. The *x*-axis represents the source dataset companies

As shown in Figure 7, for the same neural network, as more companies' data are used in pretraining, the advantage of our pretrain-then-fine-tune approach becomes more apparent, though it eventually stabilizes when pretraining with data from four or more companies. This shows that an important consideration for maximizing the benefits of our pretrain-then-fine-tune approach is the diversity of pretraining data. Moreover, since the advantage of our approach when pretraining with data from company 1, 2, 3, 6 is similar to that when pretraining with data from company 1, 2, 3, 4, 5, 6, the average increasement is convergent when the source dataset is large and diverse enough. It also implies that there may be an optimal number of companies or dataset size to establish a good enough pretrained model, beyond which increasing data diversity results in diminishing marginal performance improvement.

# 5 Discussion

## 5.1 Academic contributions

In this study, we introduced a transfer learning strategy, the pretrain-then-fine-tune approach to safety analytics and, in particular, accident prediction with safety climate data. We also proposed SafeNet, a deep neural network with residual connections to fit the tabular safety climate dataset. Moreover, we developed three evaluation indicators to compare the performance between models trained from scratch and models developed through the pretrain-then-fine-tune approach. In summary, our study enriches safety research methods and contributes to the emerging "AI for safety" research field (Kurzidem et al., 2023).

## 5.2 Insights for the trucking industry

This study also presents many contributions to the trucking industry. First, it demonstrates the feasibility of using a transfer learning approach for companies with limited data to use models pretrained on other companies' data. In practice, this means that relevant industrial associations, unions or guilds may pool data from companies and develop pretrained models for their member companies' use, if not public use. This sidesteps the privacy concerns of sharing safety data in the public domain (Hébert et al., 2021) since the relevant association may make available only the AI model but not the raw data. This facilitates and encourages the sharing of knowledge and resources in the industry on such lifesaving matters. Second, and related to the above, results of this study show the benefits of pretraining the AI model with a diverse set of data. This means the relevant associations should forge alliances across big and small companies in the industry so that they all contribute to the creation of such pretrained models by providing diverse sources of data. Third, results of this study also show that when using transfer learning for accident prediction, even though the user may directly use models pretrained with other companies' data without modification, one may still wish to fine-tune the model with one's own company's data for better accuracy. The pretrain-then-fine-tune approach proposed in this study presents a transfer learning practice that companies may emulate to derive maximal benefits from their own data and the industrial resources when the relevant pretrained models become available.

## 5.3 Insights on the application of transfer learning

Findings from this study present a few practical insights on the application of transfer learning in safety analytics.. First, to achieve better performance after fine-tuning, pretrained models with higher accuracy should be selected. Conversely, repeatedly obtaining a low accuracy of the pretrained model may suggest that the model parameters or the algorithm are not suited to the data at hand such that the benefit of pretraining is limited. Second, pretraining on smaller datasets may still be beneficial for analyzing bigger datasets compared to doing so without pretraining, especially for tabular datasets. This means bigger companies also stand to benefit from data contributions from smaller companies. Third, pretraining with data collected from more companies improves the fine-tuning accuracy. This points to the importance of data diversity to transfer learning rather than just size. That said, the diminishing marginal return of increasing pretraining data diversity in transfer learning explained in section 4.3 deserves further research.

At a more general level, the study highlights the importance of the accumulation and publication of industry-level datasets and models. More public datasets benefit the advancement of safety analytics technology, riding on the recent academic interest in AI for Safety. Such datasets help companies and researchers develop their AI models and result in benefits to both workers' well-being and economic productivity. Moreover, measures can be taken to address or avoid data privacy and confidentiality concerns, as explained earlier. For example, if one cannot share private datasets with international researchers due to data confidentiality requirements, the sharing of pretrained models is less likely to breach confidentiality. We encourage stakeholders with industrial data to consider the option of sharing pretrained models with researchers to promote knowledge transfer and technological advancement in the industry. Another possible strategy to the constraints on data sharing is to use AI-generated content (AIGC) and share the synthetic datasets from AIGC models. This is an increasingly promising option given the rapid development of generative AI like ChatGPT-4 and Llama. The datasets may be synthesized to capture important traits of the original private datasets and be made available to companies and researchers, thus promoting technological advancement globally.

In summary, this study aims to provide companies and researchers useful advice and methods so that they can develop their models in different safety analytics tasks. To this end, we have explained our proposed methods and insights derived from the experiments in detail. Consistent with our call for the sharing of data and pretrained models for industrial and academic use, we make our code publicly available at https://github.com/NUS-DBE/Pretrain-Finetune-safety-climate/ where we explain how we coded and trained the AI models for this study.

## 5.4 Limitations

In this study, we only used truck drivers' safety climate perceptions as features, so one should exercise caution when generalizing the results to other accident prediction scenarios. Future studies may investigate the suitability of our proposed transfer learning approach to accident prediction using project-related or equipment telemetry data.

## 6 Conclusions

Safety analytics are important tools to prevent WSH incidents. Unlike CV and NLP, there are limited publicly available datasets to train large-scale AI safety analytics models due to the sensitivity of data. In general, most safety analytics datasets cannot be shared with global researchers and companies. This study aims to help companies with small datasets to develop effective safety analytics AI models by transferring domain knowledge from companies with more data.

We propose a pretrain-then-fine-tune transfer learning approach aiming to help any target company leverage other companies' data for more accurate predictions of accident risks using AI models. We also developed SafeNet, a DNN algorithm for classification tasks suitable for accident prediction. Using the safety climate survey data from seven trucking companies, we show that our proposed approach results in better model performance compared to training the model from scratch using a target company's data only. We also show that in order for the transfer learning model to be effective, the pretrained model should be developed with data from diverse sources and have a high accuracy. The trucking industry may, thus, consider pooling safety analytics data from a wide range of companies and share the pretrained model using it for better knowledge and resource transfer within the industry. Through the above findings, we aim to provide insights to guide companies and researchers in developing their models on different safety analytics tasks. We also release our overall codes at https://github.com/NUS-DBE/Pretrain-Finetune-safety-climate so that the latest technology can be leveraged to make the industry safer and more sustainable.

## 7 Acknowledgement

The data in this study were collected while one of the authors worked at Liberty Mutual Research Institute for Safety. We thank the following team members for their invaluable assistance: Michelle Robertson, Susan Jeffries, Peg Rothwell, and Angela Garabet for data collection, analysis, and general assistance. This research did not receive any specific grant from funding agencies in the public, commercial, or not-for-profit sectors.

## 8 References

Argyriou, A., Evgeniou, T., & Pontil, M. (2006). Multi-task feature learning. *Proceedings of the 19th International Conference on Neural Information Processing Systems*, 41–48.

Chawla, N. V., Bowyer, K. W., Hall, L. O., & Kegelmeyer, W. P. (2002). SMOTE: Synthetic Minority Over-sampling Technique. *Journal of Artificial Intelligence Research*, *16*, 321–357. https://doi.org/10.1613/jair.953

Christian, M. S., Bradley, J. C., Wallace, J. C., & Burke, M. J. (2009). Workplace safety: A meta-analysis of the roles of person and situation factors. *Journal of Applied Psychology*, *94*(5), 1103–1127. https://doi.org/10.1037/a0016172

Dai, W., Yang, Q., Xue, G.-R., & Yu, Y. (2007). Boosting for transfer learning. *Proceedings of the 24th International Conference on Machine Learning*, 193–200. https://doi.org/10.1145/1273496.1273521

Devlin, J., Chang, M.-W., Lee, K., & Toutanova, K. (2018, October 11). *BERT: Pre-training of Deep Bidirectional Transformers for Language Understanding*. arXiv.Org. https://arxiv.org/abs/1810.04805v2

Dos Santos, E. M., Sabourin, R., & Maupin, P. (2009). Overfitting cautious selection of classifier ensembles with genetic algorithms. *Information Fusion*, *10*(2), 150–162. https://doi.org/10.1016/j.inffus.2008.11.003

Goldberg, D. M. (2022). Characterizing accident narratives with word embeddings: Improving accuracy, richness, and generalizability. *Journal of Safety Research*, *80*, 441–455. https://doi.org/10.1016/j.jsr.2021.12.024

He, Y., Huang, Y.H., Lee, J., Lytle, B., Asmone, A. S., & Goh, Y. M. (2022). A mixed-methods approach to examining safety climate among truck drivers. *Accident Analysis & Prevention*, *164*, 106458. https://doi.org/10.1016/j.aap.2021.106458

Hébert, A., Marineau, I., Gervais, G., Glatard, T., & Jaumard, B. (2021). Can we Estimate Truck Accident Risk from Telemetric Data using Machine Learning? *2021 IEEE International Conference on Big Data (Big Data)*, 1827–1836. https://doi.org/10.1109/BigData52589.2021.9671967

Houlsby, N., Giurgiu, A., Jastrzebski, S., Morrone, B., De Laroussilhe, Q., Gesmundo, A., Attariyan, M., & Gelly, S. (2019). *Parameter-efficient transfer learning for NLP*. 2790–2799.

Huang, Y.H, He, Y., Lee, J., & Hu, C. (2021). Key drivers of trucking safety climate from the perspective of leader-member exchange: Bayesian network predictive modeling approach. *Accident; Analysis and Prevention*, *150*, 105850. https://doi.org/10.1016/j.aap.2020.105850

Huang, Y.H, Lee, J., McFadden, A. C., Murphy, L. A., Robertson, M. M., Cheung, J. H., & Zohar, D. (2016). Beyond safety outcomes: An investigation of the impact of safety climate on job satisfaction, employee engagement and turnover using social exchange theory as the theoretical framework. *Applied Ergonomics*, *55*, 248–257. https://doi.org/10.1016/j.apergo.2015.10.007

Huang, Y.H, Lee, J., McFadden, A. C., Rineer, J., & Robertson, M. M. (2017). Individual employee's perceptions of "Group-level Safety Climate" (supervisor referenced) versus "Organization-level Safety Climate" (top management referenced): Associations with safety outcomes for lone workers. *Accident Analysis & Prevention*, *98*, 37–45. https://doi.org/10.1016/j.aap.2016.09.016

Huang, Y.H., Zohar, D., Robertson, M. M., Garabet, A., Lee, J., & Murphy, L. A. (2013). Development and validation of safety climate scales for lone workers using truck drivers as exemplar. *Transportation Research Part F: Traffic Psychology and Behaviour*, *17*, 5–19. https://doi.org/10.1016/j.trf.2012.08.011

Hung, P. D., & Su, N. T. (2021). Unsafe Construction Behavior Classification Using Deep Convolutional Neural Network. *Pattern Recognition and Image Analysis*, *31*(2), 271–284. https://doi.org/10.1134/S1054661821020073

Insurance Institute for Highway Safety. (2023, May). *Fatality Facts 2021: Large trucks*. IIHS-HLDI Crash Testing and Highway Safety. https://www.iihs.org/topics/fatality-statistics/detail/large-trucks

Kao, K.-Y., Thomas, C. L., Spitzmueller, C., & Huang, Y.H. (2021). Being present in enhancing safety: Examining the effects of workplace mindfulness, safety behaviors, and safety climate on safety outcomes. *Journal of Business and Psychology*, *36*(1), 1–15. https://doi.org/10.1007/s10869-019-09658-3

Kim, H., Kim, H., Hong, Y. W., & Byun, H. (2018). Detecting Construction Equipment Using a Region-Based Fully Convolutional Network and Transfer Learning. *Journal of Computing in Civil Engineering*, *32*(2), 04017082. https://doi.org/10.1061/(ASCE)CP.1943-5487.0000731

Kirillov, A., Mintun, E., Ravi, N., Mao, H., Rolland, C., Gustafson, L., Xiao, T., Whitehead, S., Berg, A. C., Lo, W.-Y., Dollár, P., & Girshick, R. (2023, April 5). *Segment Anything*. arXiv.Org. https://arxiv.org/abs/2304.02643v1

Kohavi, R., & Sommerfield, D. (1995, August 20). *Feature Subset Selection Using the Wrapper Method: Overfitting and Dynamic Search Space Topology*. Knowledge Discovery and Data Mining.

Kurzidem, I., Burton, S., & Schleiss, P. (2023). AI for Safety: How to use Explainable Machine Learning Approaches for Safety Analyses. *CEUR Workshop Proceedings*. The IJCAI-2023 AISafety and SafeRL Joint Workshop, Macao.

Lawrence, N. D., & Platt, J. C. (2004). Learning to learn with the informative vector machine. *Proceedings of the Twenty-First International Conference on Machine Learning*, 65. https://doi.org/10.1145/1015330.1015382

Lee, J., Huang, Y.H., Sinclair, R. R., & Cheung, J. H. (2019). Outcomes of Safety Climate in Trucking: A Longitudinal Framework. *Journal of Business and Psychology*, *34*(6), 865–878. https://doi.org/10.1007/s10869-018-9610-5

Lee, J., & Lee, S. (2023). Construction Site Safety Management: A Computer Vision and Deep Learning Approach. *Sensors*, *23*(2), Article 2. https://doi.org/10.3390/s23020944

Leoni, L., BahooToroody, A., Abaei, M. M., Cantini, A., BahooToroody, F., & De Carlo, F. (2024). Machine learning and deep learning for safety applications: Investigating the intellectual structure and the temporal evolution. *Safety Science*, *170*, 106363. https://doi.org/10.1016/j.ssci.2023.106363

Lynch, S. M. (2007). *Introduction to applied Bayesian statistics and estimation for social scientists*. Springer. https://go.exlibris.link/Vxh2T5dQ

Mihalkova, L., Huynh, T., & Mooney, R. J. (2007). Mapping and revising Markov logic networks for transfer learning. *Proceedings of the 22nd National Conference on Artificial Intelligence - Volume 1*, 608–614.

Ministry of Manpower. (2023). *Workplace Safety and Health Report 2023*. Ministry of Manpower.

Mutegeki, R., & Han, D. S. (2019). Feature-Representation Transfer Learning for Human Activity Recognition. *2019 International Conference on Information and Communication Technology Convergence (ICTC)*, 18–20. https://doi.org/10.1109/ICTC46691.2019.8939979

Nahrgang, J. D., Morgeson, F. P., & Hofmann, D. A. (2011). Safety at work: A meta-analytic investigation of the link between job demands, job resources, burnout, engagement, and safety outcomes. *The Journal of Applied Psychology*, *96*(1), 71–94. https://doi.org/10.1037/a0021484

National Safety Council. (2023). Large Trucks. *Injury Facts*. https://injuryfacts.nsc.org/motor-vehicle/road-users/large-trucks/

Niu, S., Liu, Y., Wang, J., & Song, H. (2020). A Decade Survey of Transfer Learning (2010–2020). *IEEE Transactions on Artificial Intelligence*, *1*, 151–166. https://doi.org/10.1109/TAI.2021.3054609

Pan, S. J., & Yang, Q. (2009). A survey on transfer learning. *IEEE Transactions on Knowledge and Data Engineering*, *22*(10), 1345–1359.

Poh, C. Q. X., Ubeynarayana, C. U., & Goh, Y. M. (2018). Safety leading indicators for construction sites: A machine learning approach. *Automation in Construction*, *93*, 375–386. https://doi.org/10.1016/j.autcon.2018.03.022

Rajput, D., Wang, W.-J., & Chen, C.-C. (2023). Evaluation of a decided sample size in machine learning applications. *BMC Bioinformatics*, *24*(1), 48. https://doi.org/10.1186/s12859-023-05156-9

Rusu, A. A., Rabinowitz, N. C., Desjardins, G., Soyer, H., Kirkpatrick, J., Kavukcuoglu, K., Pascanu, R., & Hadsell, R. (2016). Progressive neural networks. *arXiv Preprint arXiv:1606.04671*.

Sarkar, S., & Maiti, J. (2020). Machine learning in occupational accident analysis: A review using science mapping approach with citation network analysis. *Safety Science*, *131*, 104900. https://doi.org/10.1016/j.ssci.2020.104900

Sun, K., Lan, T., Goh, Y. M., Safiena, S., Huang, Y.-H., Lytle, B., & He, Y. (2024). An interpretable clustering approach to safety climate analysis: Examining driver group distinctions. *Accident Analysis and Prevention*, *196*, 107420. https://doi.org/10.1016/j.aap.2023.107420

Tang, J. C., Ab. Nasir, A. F., P. P. Abdul Majeed, A., Mohd Razman, M. A., Mohd Khairuddin, I., & Lim, T. L. (2022). Vision-Based Human Presence Detection by Means of Transfer Learning Approach. In I. Mohd. Khairuddin, M. A. Abdullah, A. F. Ab. Nasir, J. A. Mat Jizat, Mohd. A. Mohd. Razman, A. S. Abdul Ghani, M. A. Zakaria, W. H. Mohd. Isa, & A. P. P. Abdul Majeed (Eds.), *Enabling Industry 4.0 through Advances in Mechatronics* (pp. 571–580). Springer Nature. https://doi.org/10.1007/978-981-19-2095-0_49

Wang, Y., Zhao, P., & Zhang, Z. (2023). A deep learning approach using attention mechanism and transfer learning for electromyographic hand gesture estimation. *Expert Systems with Applications*, *234*, 121055. https://doi.org/10.1016/j.eswa.2023.121055

Yin, X., Yu, X., Sohn, K., Liu, X., & Chandraker, M. (2019). Feature Transfer Learning for Face Recognition With Under-Represented Data. *2019 IEEE/CVF Conference on Computer Vision and Pattern Recognition (CVPR)*, 5697–5706. https://doi.org/10.1109/CVPR.2019.00585

Zhao, Z., Alzubaidi, L., Zhang, J., Duan, Y., & Gu, Y. (2024). A comparison review of transfer learning and self-supervised learning: Definitions, applications, advantages and limitations. *Expert Systems with Applications*, *242*, 122807. https://doi.org/10.1016/j.eswa.2023.122807

Zhuang, F., Qi, Z., Duan, K., Xi, D., Zhu, Y., Zhu, H., Xiong, H., & He, Q. (2021). A Comprehensive Survey on Transfer Learning. *Proceedings of the IEEE*, *109*(1), 43–76. https://doi.org/10.1109/JPROC.2020.3004555

Zohar, D. (1980). Safety climate in industrial organizations: Theoretical and applied implications. *Journal of Applied Psychology*, *65*(1), 96–102. https://doi.org/10.1037/0021-9010.65.1.96

Zohar, D., & Luria, G. (2005). A Multilevel Model of Safety Climate: Cross-Level Relationships Between Organization and Group-Level Climates. *Journal of Applied Psychology*, *90*(4), 616–628. https://doi.org/10.1037/0021-9010.90.4.616